\documentclass[runningheads]{llncs}
\usepackage{graphicx}
\usepackage{subcaption}
\usepackage{caption}
\usepackage{multirow}
\usepackage{array}
\usepackage{ulem}
\usepackage{booktabs}
\usepackage{tabularx}

\begin{document}
\title{Predicting decision-making in the future: \\Human versus Machine\vspace{-0.4cm}}
\titlerunning{Predicting decision-making in the future: Human vs Machine}

\author{Hoe Sung Ryu\inst{1}\orcidID{0000-0002-9515-4402} \newline
Uijong Ju \inst{2}\orcidID{0000-0002-9391-3938} \newline
Christian Wallraven\inst{3,*}\orcidID{0000-0002-2604-9115}}
\authorrunning{Ryu. et al.}

\institute{Department of Artificial Intelligence, Korea University, Seoul, Korea \inst{1} \newline
\email{hoesungryu@korea.ac.kr}\newline
Department of Information Display, Kyung Hee University, Seoul, Korea \inst{2} \newline
\email{juuijong@khu.ac.kr}\newline
Department of Artificial Intelligence \& Department of Brain and Cognitive Engineering, Korea University, Seoul, Korea \inst{3} \newline
\email{wallraven@korea.ac.kr}
}

\maketitle
\vspace{-0.7cm}
\begin{abstract}
Deep neural networks (DNNs) have become remarkably successful in data prediction, and have even been used to predict future actions based on limited input. This raises the question: do these systems actually ``understand" the event similar to humans? Here, we address this issue using videos taken from an accident situation in a driving simulation. In this situation, drivers had to choose between crashing into a suddenly-appeared obstacle or steering their car off a previously indicated cliff. We compared how well humans and a DNN predicted this decision as a function of time before the event. The DNN outperformed humans for early time-points, but had an equal performance for later time-points. Interestingly, spatio-temporal image manipulations and Grad-CAM visualizations uncovered some expected behavior, but also highlighted potential differences in temporal processing for the DNN. \vspace{-0.5cm}

\keywords{Deep Learning \and Video Prediction \and Humans versus Machines \and Decision-Making \and Video Analysis} 
\end{abstract}
\vspace{-0.95cm}

\section{Introduction}
\vspace{-0.2cm}
The ability to predict, anticipate and reason about future events is the essence of intelligence \cite{hawkins2004intelligence} and one of the main goals of decision-making systems \cite{edwards1954theory}. In general, predicting human behavior in future situations is of course an extremely hard task in an unconstrained setting. Given additional information about the context and the type of behavior to be predicted, however, behavior forecasting becomes a more tractable problem.

Recently, deep neural networks (DNNs) have dramatically improved the state-of-the-art in speech recognition, visual object recognition, object detection, and many other domains \cite{goodfellow2016deep}. Although DNNs yield excellent performance in such applications, it is often difficult to get insights into how and why a certain classification result has been made. To address this issue, Explainable Artificial Intelligence (XAI) proposes to make a shift towards more transparent AI \cite{adadi2018peeking}. As part of this shift, research has focused on comparing DNN performance in a task with human performance to better understand the underlying decision-making capacities of DNNs \cite{tan2019efficientnet}. Here it becomes important to look closely at the metric with which performance is measured - for example, if humans and DNNs have the same, high accuracy in identifying COVID-19 chest radiographs, this does not mean that DNNs use the same image-related cues to solve the task \cite{degrave2020ai}. Therefore, \cite{oprea2020review} proposed to use metrics \textit{beyond accuracy} to understand more deeply how DNNs and humans differ. Methods from XAI, such as feature visualization, for example, can help to understand which visual input is important for the network in making a certain decision \cite{degrave2020ai}. 

Examples of behavior prediction that have been tackled with DNNs recently include, for example, predicting a future action by observing only a few portions of an action in progress \cite{rodriguez2018action}, anticipating the next word in textual data given context \cite{mikolov2010recurrent}, or predicting human driving behavior before dangerous situations occur \cite{ontanon2017learning}. Hence, DNNs seem to be capable of analyzing the spatio-temporal contents of an event to predict an outcome - the important question, then, becomes, do these networks actually ``understand" the event similar to humans, or do they use spurious, high-dimensional correlations to form their prediction \cite{degrave2020ai}?

In the present work, we take an accident situation during driving as a challenging context for studying this question. To explain the situation, imagine you are driving and there is a fork ahead; a prior warning sign alerted you that one of the directions of the fork will lead to a cliff, which will fatally crash the car. The other direction seems safe to drive, until, suddenly an obstacle appears on this direction of the fork, blocking the safe path. How do you react to this sudden change of circumstance? Understanding and modeling human behavior and its underlying factors in such situations can teach us a lot about decision-making under pressure and with high-risk stakes and has many important application areas.

Since it is impossible to study such a situation in the real world, a recent study by \cite{ju2020acoustic} employed virtual reality (VR) to investigate exactly this event. In this experiment, participants were trained to navigate a driving course that contained multiple, warning-indicated forks. The aforementioned accident situation was inserted only during the final test-run to see how participants would react to the sudden appearance of the obstacle (turn left and crash into the obstacle, or turn right and crash the car fatally off the cliff). From this study, here, we take the in-car videos that lead up to that final decision and segment them into time periods for predicting the turn direction. Importantly, the resulting short video segments were analyzed by \textit{both} human participants and a DNN to predict the final decision. Using this strategy, we can compare human and DNN performance in predicting decision-making, but also look closer into the decision features of the DNN, using the tools of explainable AI.\vspace{-0.3cm}
%
%
%
\section{Related Work}
\vspace{-0.2cm}
\subsection{Predicting the ``future"}
There has been growing interest in predicting the future through DNNs where machines have to react to human actions as early as possible such as autonomous driving, human-robotic interaction. In general, anticipating actions before these begin is a challenging problem, as this requires extensive contextual knowledge. 

To solve this problem, approaches have used predefined target classes and a few portions of an action from short video segments leading up to an event (e.g., \cite{rodriguez2018action} for predicting the future motion of a person). Similarly, in the field of natural language processing, \cite{mikolov2010recurrent} proposed a method of predicting what a person will say next from given contextual data.

Other methods for human trajectory prediction, include, for example, \cite{zhang2020stinet,bhattacharyya2018long} trying to predict pedestrian trajectories from 3D coordinates obtained from stereo cameras and LIDAR sensors using deep learning-based models. Although adding information from various sensors improved the prediction performance, obtaining such data is still a challenging problem in general, leading to image-only approaches such as \cite{poibrenski2021multimodal}.
\subsection{Comparisons of humans and DNNs} 
In order to better peek into the black box of DNNs, comparing performance between humans and DNNs has become an important research topic. Focusing on the human visual system, research has discussed human–machine comparisons at a conceptual level \cite{majaj2018deep,han2020scale,cichy2019deep}. Indeed, these works show that deep learning systems not only solve image classification, but also reproduce certain aspects of human perception and cognition. This even goes as far as being able to reproduce phenomena of human vision, such as illusions \cite{gomez2019convolutional} or crowding \cite{volokitin2017deep}.

In addition, several studies have been conducted trying to impart more higher-level decision-making skills into DNNs: in \cite{barrett2018measuring}, for example, ResNet models were able to solve abstract visual reasoning problems (IQ test) significantly better than humans - it is highly unlikely, however, that the model’s knowledge representation matches that of humans, as the resulting networks were not able to learn visual relationships efficiently and robustly \cite{funke2021five} - see also the many examples of adversarial attacks \cite{akhtar2018threat}.

In other studies, input stimuli are manipulated or degraded to determine important visual features for human or machine decision-making. In \cite{geirhos2018generalisation}, for example, twelve different types of image degradation were tested to compare error-patterns between humans and DNNs in classification. The authors found that the human visual system seemed to be more robust for image manipulations, but it DNNs trained directly on the degraded images were able to outperform humans. Similarly, based on adversarial perturbations, \cite{cichy2019deep,ritter2017cognitive} indicate that DNNs may have the potential ability to provide predictions and explanations of cognitive phenomena.

Overall, the field therefore seems to have quite heterogenous results when trying to compare human and machine performance with some studies finding commonalities and others finding critical differences. Here, we try to add to this general topic by comparing human and DNN performance in predicting a decision in a critical accident situation and analyzing the critical DNN features for prediction using similar tools.
\section{Method}
\subsection{Experiment setup}
\begin{figure}[t!]
     \centering
     \begin{subfigure}[b]{0.48\textwidth}
         \centering
         \includegraphics[width=\textwidth]{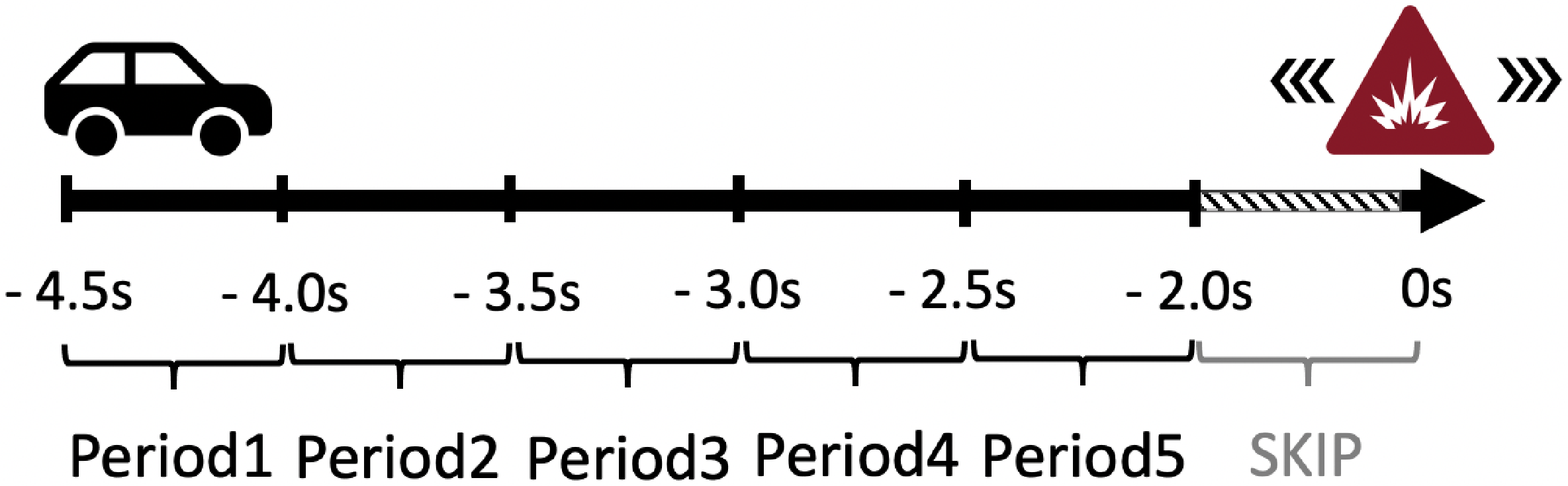}
         \caption{Flow chart of the segmentation}
         \label{subfig:figure_trimmed_video}
     \end{subfigure}
     \begin{subfigure}[b]{0.48\textwidth}
         \centering
         \includegraphics[width=\textwidth]{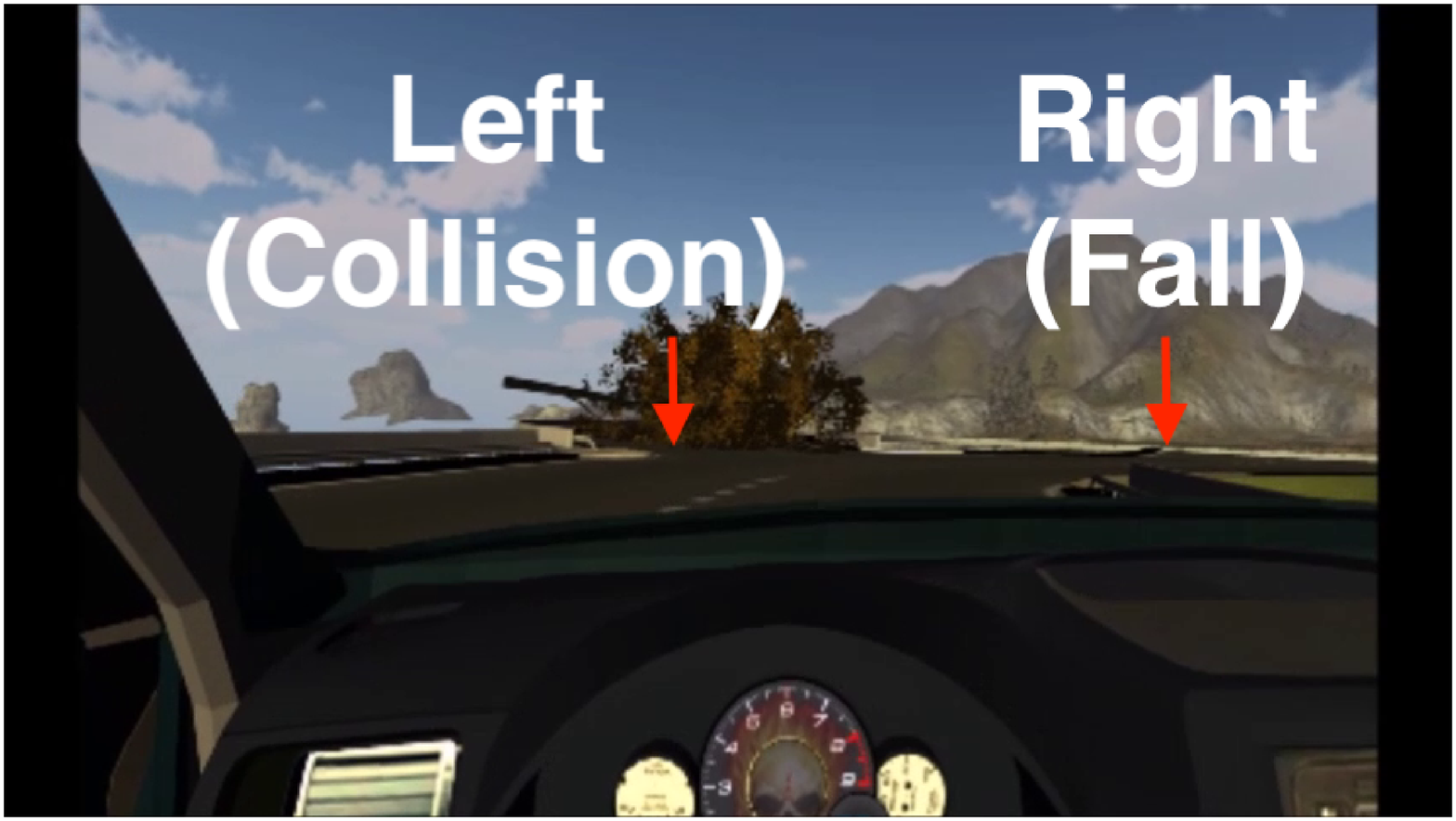}
         \caption{Final frame in the segmented video}
         \label{subfig:figure_description}
     \end{subfigure}
        \caption{(a) Segmentation setup and (b) screenshot from experiment video}
        \label{fig:Experiment_setup}
\vspace{-0.5cm}
\end{figure}
\subsubsection{Data:} 
To compare the prediction of decision-making between humans and DNNs, we used the original, in-car videos from \cite{ju2020acoustic} - both humans and DNNs received the same data and the same task for a fair comparison. First, we trimmed the total of $n_{\mathrm{total}}=74$ videos from all participants to $2$ to $4.5$ seconds before the final decision - hence, the actual decision in which the car turned left or right was not part of the video. Let $D$ denote our set of trimmed videos. Second, we partitioned $D$ into five non-overlapping subsets $D_p \subset \{{D_{1}, D_{2}, D_{3}, D_{4}, D_{5} } \}$ with each $D_{p}$ being a $0.5$s-long video, containing $16$ frames at $480$px$\times720$px (width$\times$height). For each video, this yields a total of $5$ time segments such that, for example, $D_{5}$ indicates the final segment running from $-2.5$s to $-2.0$s, where $0$s would indicate the actual time of the decision. In addition, each of the $D_p$ is labelled either as fall ($n_{\mathrm{fall}}=23$) and collision ($n_{\mathrm{collide}}=51$). Given the human decision proportions in this accident situation (most people chose to collide with the obstacle, rather than to crash their car off the cliff), this results in an imbalanced, binary dataset with label ratios of $\approx31\%$ versus $\approx69\%$, respectively.
\subsection{Behavioral experiment}
\subsubsection{Participants:}
A total of $29$ participants ($18$ females, mean age $24.69\pm 3.11$(SD)) were recruited from the student population of Korea University. All participants had normal or corrected-to-normal vision and possessed a driver’s license. The experiment adhered to the tenets of the Declaration of Helsinki and was approved by the Institutional Review Board of Korea University with IRB number KUIRB-2018-0096-02s.

\subsubsection{Experimental procedure:}
As for the experimental procedure, participants were tested in a small, enclosed room with no distractions. Upon entering the room, the procedure of the experiment was explained to them - in particular, that the video clips were part of a longer video sequence leading up to a final decision by a driver whether to collide with the trees or to fall down the cliff. Participants were not informed that the decision ratio of the dataset was imbalanced. The whole experiment was conducted on a laptop running at $480$px$\times 720$px resolution - participants sat a distance of $\approx60$ centimeters, with video segments subtending a visual angle of $\approx30^{\circ}$. The behavioral experiment was created in PsychoPy (v3.0) \cite{peirce2019psychopy2}. 

Each experiment had a sequence of $370 = 5\times 74$ trials, in which participants were asked to determine whether the car was going to collide or fall. In each trial, a short video segment $D_{p}$ was randomly chosen and repeatedly shown to the participant until they felt they knew the answer, at which time they were to press the space bar. There was no time limit set by the experimenter, nor were participants explicitly instructed to respond as quickly or as accurately as possible. After the space bar was pressed, the time from start of the segment to the key press was recorded as response time, the video segment stopped looping and disappeared, and a text appeared in the center of the screen: ``Which direction will the car go: collide or fall?''. Participants were to press the right arrow button for a collision and the left arrow button for a fall decision. All video segments were pseudo-randomly chosen and shown only once. Dependent variables were response time and response.
\subsection{Computational experiment}
\begin{table}[t!]
\begin{center}
\begin{tabular}{c|c|c}
\hline
\multicolumn{1}{c|}{Layer Name} & \multicolumn{1}{c|}{Filter Shape} & \multicolumn{1}{c}{Repeats} \\ \hline
										conv1            & $7 \times 7 \times 7 \times$, $64$, stride $1$                   & 1  \\ \hline 
\multirow{3}{*}{conv$2\_$x}         & $3\times3\times3$ max pool stride $2$           & \multirow{3}{*}{2}                     \\
                                 & $ \left[ \begin{array}{cc}
                                    3 \times 3 \times 3, & 64\\
                                    3 \times 3 \times 3, & 64
                                     \end{array}\right]$ \\  \hline
\multirow{2}{*}{conv$3\_$x}         &  $ \left[ \begin{array}{cc}
                                    3 \times 3 \times 3, & 64\\
                                    3 \times 3 \times 3, & 64
                                     \end{array}\right]$           & 2                     \\ \hline 
\multirow{2}{*}{conv$4\_$x}         & $ \left[ \begin{array}{cc}
                                    3 \times 3 \times 3, & 64\\
                                    3 \times 3 \times 3, & 64
                                     \end{array}\right]$           & 2                     \\ \hline
\multirow{2}{*}{conv$5\_$x}         &  $ \left[ \begin{array}{cc}
                                    3 \times 3 \times 3, & 64\\
                                    3 \times 3 \times 3, & 64
                                     \end{array}\right]$           & 2                   \\\hline
\multicolumn{3}{c}{Average Pooling, $512\mathrm{-}d$ FC, Sigmoid} \\ \hline    
\end{tabular}
\label{architectures}
\end{center}
\caption{ResNet(2+1)D architecture in our experiments. Convolutional residual blocks are shown illustrated in brackets, next to the number of times each block is repeated in the stack.}
\vspace{-0.9cm}
\end{table}
\subsubsection{ResNet(2+1)D architecture:}
A popular architecture in action recognition consists of a 3D convolutional neural network (CNN), which extends the typical 2D filters of image-based CNNs to 3D convolutional filters. This approach therefore directly extracts spatio-temporal features from videos by creating hierarchical representations and was shown to perform well on large-scale video datasets (e.g.,  \cite{carreira2017quo}). In our experiments, we use a similar ResNet(2+1)D architecture \cite{tran2018closer} in which the 3D filters are factorized into 2D spatial convolutions and a 1D temporal convolution, which improves optimization. 

In ResNet(2+1)D, the input tensor $D_p$ ($p \in \{1, 2,3,4,5\}$) is $5D$ and has size $B\times F\times C \times W \times H $ , where $B$ is  the number of mini-batch, $F$ is the number of frames of each video, $C$ is RGB channel, and $W$ and $H$ are the width and height of the frame, respectively.

Our Network takes $16$ clips consisting of RGB frames with the size of  $112$px $\times 112$px as an input. Each input frame is channel-normalized with mean ($0.43216$, $0.394666$, $0.37645$) and SD ($0.22803$, $0.22145$, $0.216989$). Down-sampling of the inputs is performed by $\mathrm{conv}1$, $\mathrm{conv}3\_1$, $\mathrm{conv}4\_1$, and $\mathrm{conv}5\_1$. Conv$1$ is implemented by convolutional striding of $1 \times 2 \times 2$, and the remaining convolutions  are implemented by striding of $2 \times 2 \times 2$. 
Since our overall sample size is small, a fine-tuning strategy with a pre-trained network was used. Its earlier layers remain fixed during training on our data with only the later layers of the network being optimized for prediction. In our experiments, the ResNet(2+1)D model is pre-trained on the Kinetics dataset \cite{tran2018closer} and fine-tuned on layers $\mathrm{conv}\_5$ upwards.

To illustrate the importance of the temporal context, we further experiment with three additional settings: models trained on $8$ frame clips (sampled uniformly from each segment) or $2$ frame clips (using only the first or the last two consecutive frames in the segment). 

During training, batch normalization is applied to all convolutional layers. We deploy the Adamax optimizer with a mini-batch size of $16$. Learning rates are updated by using a one-cycle scheduler \cite{smith2017cyclical} that starts with a small learning rate of $1\mathrm{e-}4$, which is increased after each mini-batch until the maximum learning rate of $8\mathrm{e-}3$. All processing was done on an Intel Xeon (Gold 5120 @2.20GHz) CPU and two NVIDIA V100 GPUs using Pytorch version $1.4.0$. 

\subsubsection{Performance measures:}
For the experiment, we repeated a $5$-fold cross-validation process $20$ times, which created $100$ folds in total. In every batch per epoch, we balanced the label of the training data set to $1:1$ using sampling with replacement - this was not applied to the test set.

\section{Experimental Results}
\begin{figure}[hbt!]
     \centering
     \begin{subfigure}[hb!]{0.48\textwidth}
         \centering
         \includegraphics[width=\textwidth,height=.4\textheight]{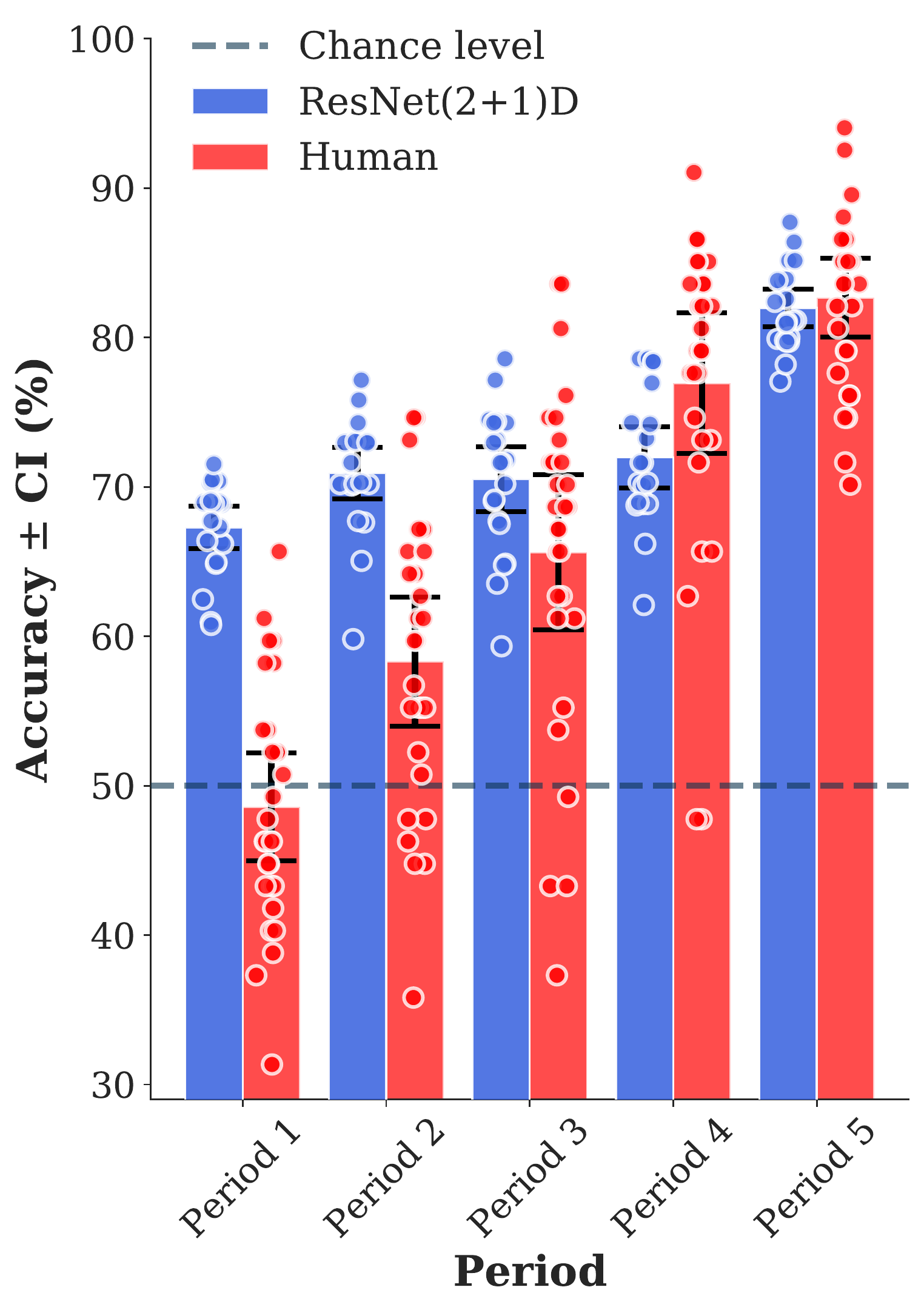}
         \caption{Prediction accuracy for humans vs DNN}
         \label{fig:figure_humanVSmachine}
     \end{subfigure}
     \begin{subfigure}[hb!]{0.48\textwidth}
         \centering
         \includegraphics[width=\textwidth,height=.3894\textheight]{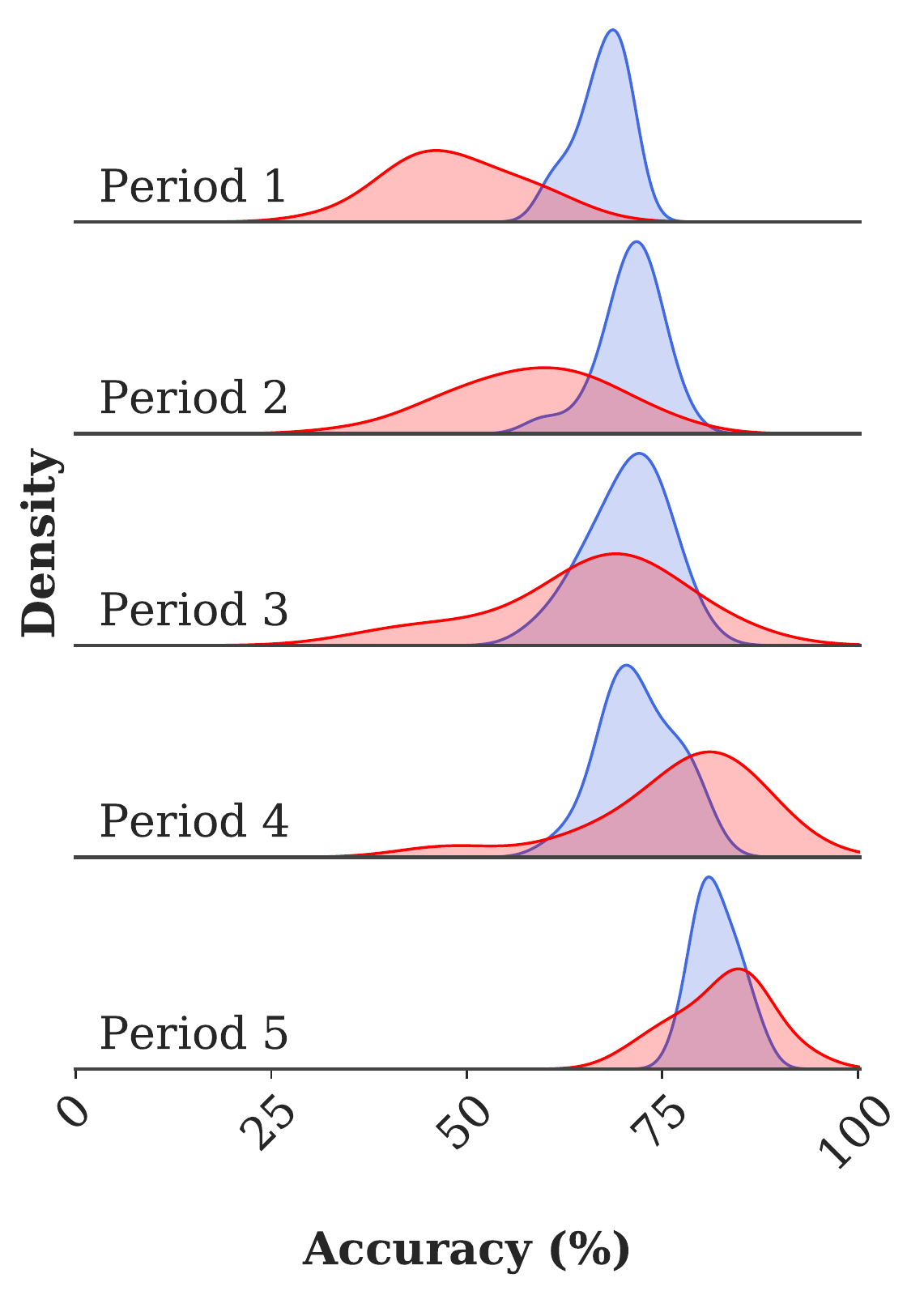}
         \caption{Kernel density estimate plot}
         \label{fig:figure_joyplot}
     \end{subfigure} 
\caption{Performance comparison of humans and ResNet(2+1)D. The margin of error was calculated at a confidence level of 95\%. Gaussian kernels were used to estimate kernel density and bandwidth calculated by Scott’s rule from \cite{scott2015multivariate}.}
\label{fig:Performance comparison of the Human and ResNet(2_1)D}
\vspace{-0.1cm}
\end{figure}
\subsection{Comparison of humans and DNN}
We compared the ResNet(2+1)D prediction results to human performance on the exact same video sequences. As Figure \ref{fig:Performance comparison of the Human and ResNet(2_1)D} shows, the model achieves higher recognition rates compared to human participants until Period 3, whereas results seemed more similar to humans in periods 4 and 5. 

A two-way analysis of variance (ANOVA) with factors of group (human or DNN) and time periods (5 periods) showed significant differences among groups ($F(1,48)=13.8882$, $p<.001$), time periods $F(4,192)=107.2769882$, $p<.001$) and the interaction between group and time periods ($F(4,192)=21.6405$, $p<.001$) - see Table \ref{table:ANO}. 

Since the interaction was significant, we next performed multiple pairwise comparisons on all possible combinations. To correct for multiple comparisons, we applied a Bonferroni correction, correcting our alpha-level to $0.005$ ($=0.05/10$). Results overall showed no significant differences in period 5 ($p=0.5853$) between human and ResNet(2+1)D, but significant differences from period 1 to 4 - see Table \ref{posthoc}.
\begin{table}[tbp!]
\begin{center}
\begin{tabular}{p{0.13\textwidth}>{\centering}p{0.13\textwidth}>{\centering}p{0.13\textwidth}>{\centering}p{0.15\textwidth}>{\centering}p{0.2\textwidth}>{\centering\arraybackslash}p{0.15\textwidth}}
   Factor    & Num DF & Den DF & F-statistic & p-value  &  $\eta^2$ \\[3pt] \hline
 Group       &     1 &    47 & 13.1187  & $< .001$ & 0.2182  \\[5pt] 
 Time        &     4 &   188 & 164.2998 & $< .001$ & 0.7775  \\[5pt] 
 Interaction &     4 &   188 & 35.9250  & $< .001$ & 0.4332  \\ \hline
\end{tabular}
\caption{Two-way ANOVA of performance comparing humans and ResNet}
\label{table:ANO}
\end{center}
\vspace{-1.3cm}
\end{table}

\begin{table}[htb!]
\begin{center}
\begin{tabular}
{p{0.18\textwidth}>{\centering}p{0.1\textwidth}>{\centering}p{0.1\textwidth}>{\centering}p{0.1\textwidth}>
{\centering}p{0.1\textwidth}>{\centering}p{0.1\textwidth}>{\centering}p{0.1\textwidth}>{\centering\arraybackslash}p{0.1\textwidth}}
Contrast      & Time    & A       & B       &        T &     dof &   p-adjust &   cohen \\[3pt] \hline
 Group        & -       & Human   & ResNet &  -4.2540  & 34.1198 &  $< .001$ & -1.0528 \\[5pt]
 Time         & -       & Period1 & Period2 &  -7.7186  & 48.0000 &  $< .001$ & -0.6803 \\[5pt]
 Time         & -       & Period1 & Period3 &  -7.2275  & 48.0000 &  $< .001$ & -1.0843 \\[5pt]
 Time         & -       & Period1 & Period4 &  -8.6652  & 48.0000 &  $< .001$ & -1.8413 \\[5pt]
 Time         & -       & Period1 & Period5 &  -14.7811 & 48.0000 &  $< .001$ & -3.0093 \\[5pt]
 Time         & -       & Period2 & Period3 &  -3.6096  & 48.0000 &  $< .001$ & -0.4241 \\[5pt]
 Time         & -       & Period2 & Period4 &  -6.8742  & 48.0000 &  $< .001$ & -1.2161 \\[5pt]
 Time         & -       & Period2 & Period5 &  -13.3034 & 48.0000 &  $< .001$ & -2.4194 \\[5pt]
 Time         & -       & Period3 & Period4 &  -5.9944	& 48.0000 &  $< .001$ & -0.7870 \\[5pt]
 Time         & -       & Period3 & Period5 &  -11.3766 & 48.0000 &  $< .001$ & -1.9326 \\[5pt]
 Time         & -       & Period4 & Period5 &  -7.0810	& 48.0000 &  $< .001$ & -1.0476 \\[5pt]
 Time * Group & Period1 & Human   & ResNet  &  -11.2901	& 38.9798 &  $< .001$ & -2.8579 \\[5pt]
 Time * Group & Period2 & Human   & ResNet  &  -6.3592	& 39.4529 &  $< .001$ & -1.6135 \\[5pt]
 Time * Group & Period3 & Human   & ResNet  &  -2.0353  & 40.1793 &  0.0484	     & -0.5183 \\[5pt]
 Time * Group & Period4 & Human   & ResNet  &   2.2670  & 40.9008 &  0.0287      &  0.5795 \\[5pt]
 Time * Group & Period5 & Human   & ResNet  &   0.5498	& 42.6486 &  0.5853      &  0.1119 \\\hline
\end{tabular}
\caption{Results of post-hoc multiple comparisons}
\label{posthoc}
\end{center}
\vspace{-1.3cm}
\end{table}
\subsection{Discriminative features - Grad-CAM attention map}

In the previous analysis, we compared only the performance of humans versus machines. Here, we employ the popular Grad-CAM method \cite{selvaraju2017grad} from explainability research to detect the most important spatial (and temporal) information for the prediction. The output of Grad-CAM is a heatmap visualization for a given class label and provides a localization map that highlights important regions in the image. 

\begin{figure}[thb!]
    \begin{subfigure}[H]{\textwidth}
    \includegraphics[width=.3\linewidth]{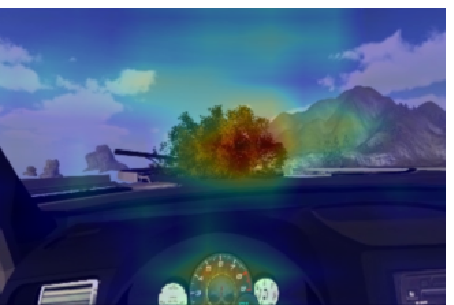}\hfill
    \includegraphics[width=.3\linewidth]{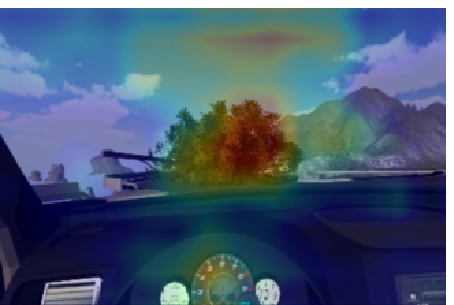}\hfill
    \includegraphics[width=.3\linewidth]{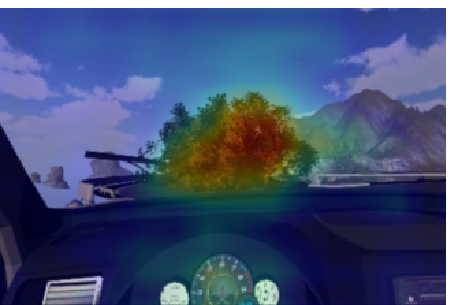}
    \caption{Turning left (collision with tree) in Period 5}
    \label{sfig:left}
    \end{subfigure}
    \\[\smallskipamount]
    \begin{subfigure}{\textwidth}
    \includegraphics[width=.3\linewidth]{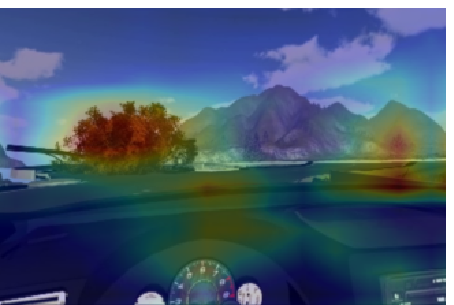}\hfill
    \includegraphics[width=.3\linewidth]{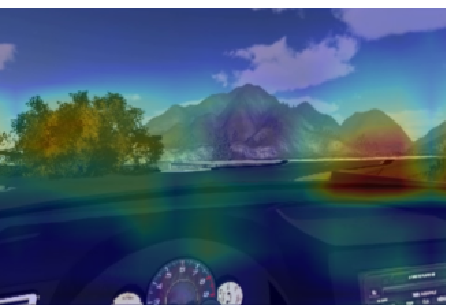}\hfill
    \includegraphics[width=.3\linewidth]{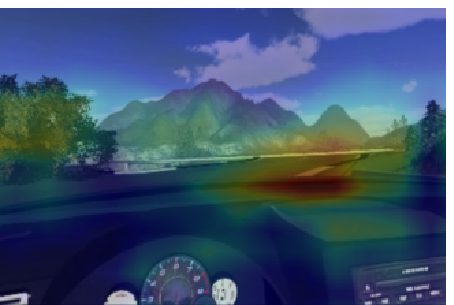}
    \caption{Turning right (fall off a cliff) in Period 5}
    \label{sfig:right}
    \end{subfigure}
    \\[\smallskipamount]
    \begin{subfigure}[H]{\textwidth}
    \includegraphics[width=.3\linewidth]{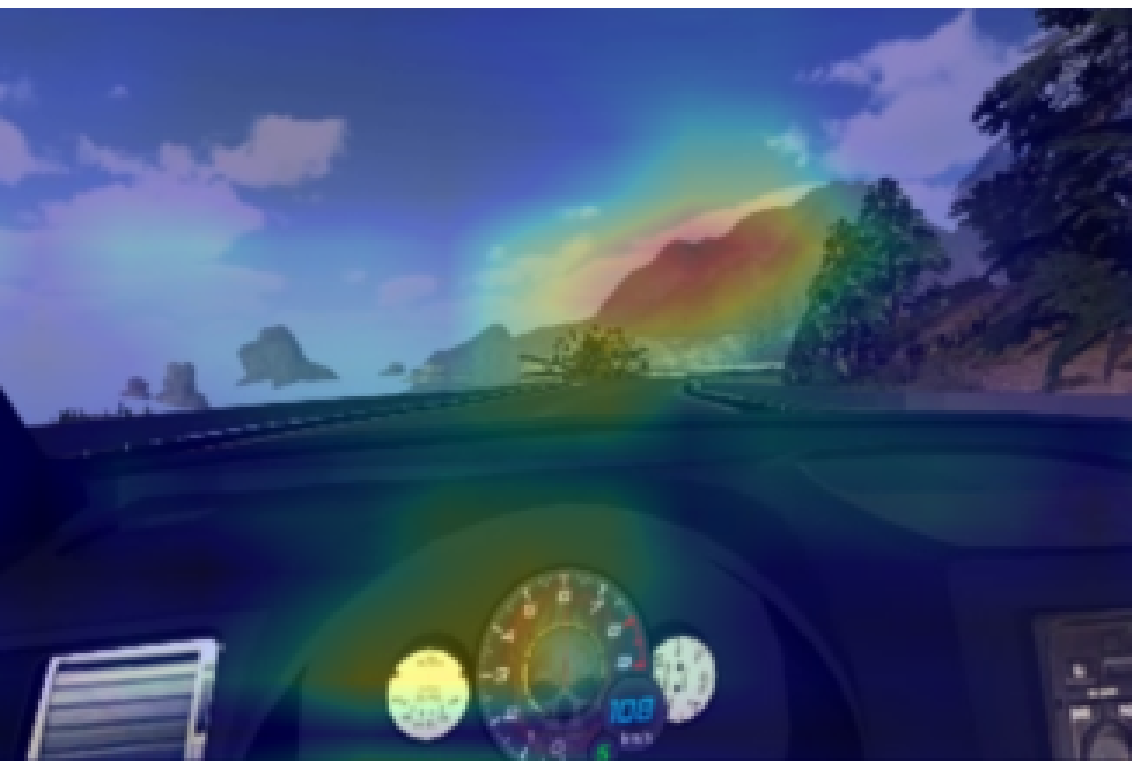}\hfill
    \includegraphics[width=.3\linewidth]{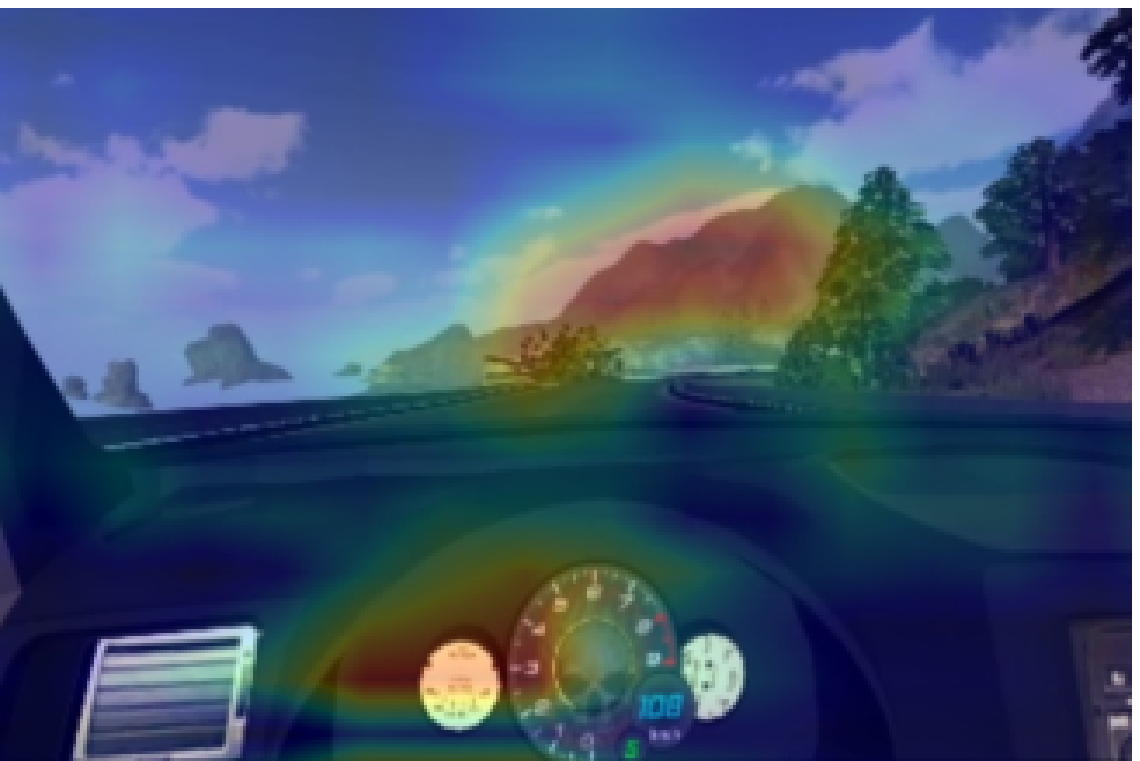}\hfill
    \includegraphics[width=.3\linewidth]{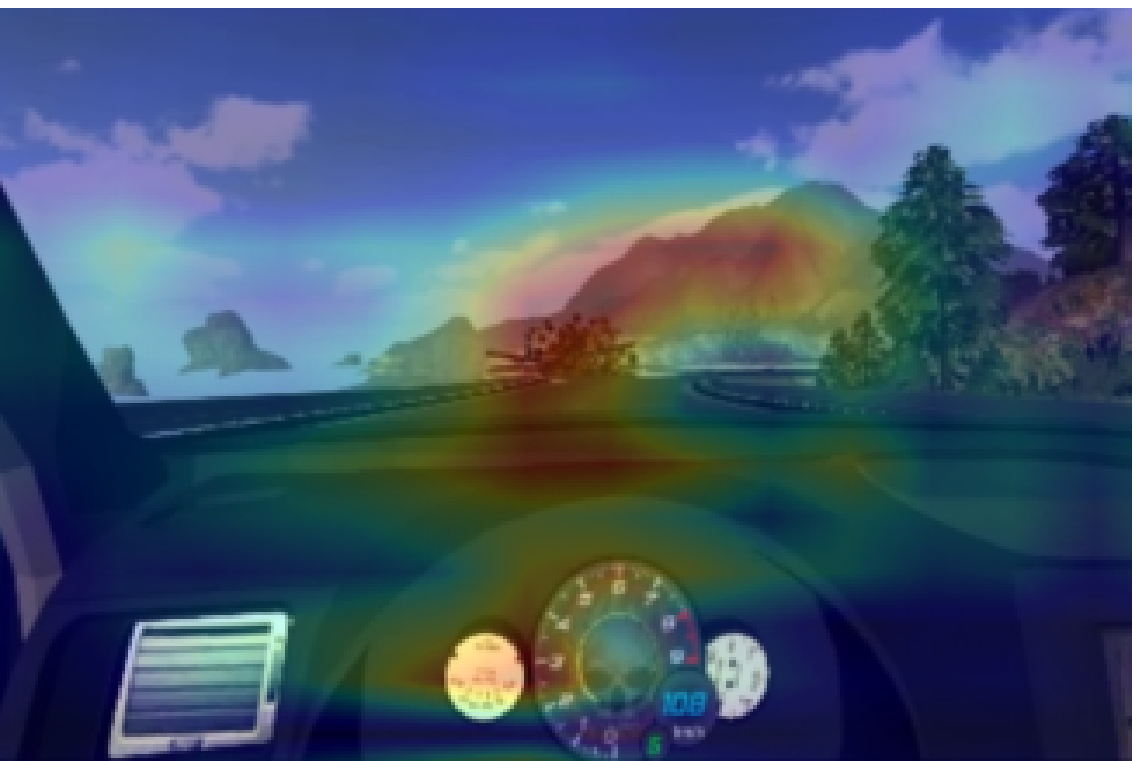}
    \caption{Turning left (collision with tree) in Period 1}
    \label{sfig:left in period1}
    \end{subfigure}
    \\[\smallskipamount]
    \begin{subfigure}{\textwidth}
    \includegraphics[width=.3\linewidth]{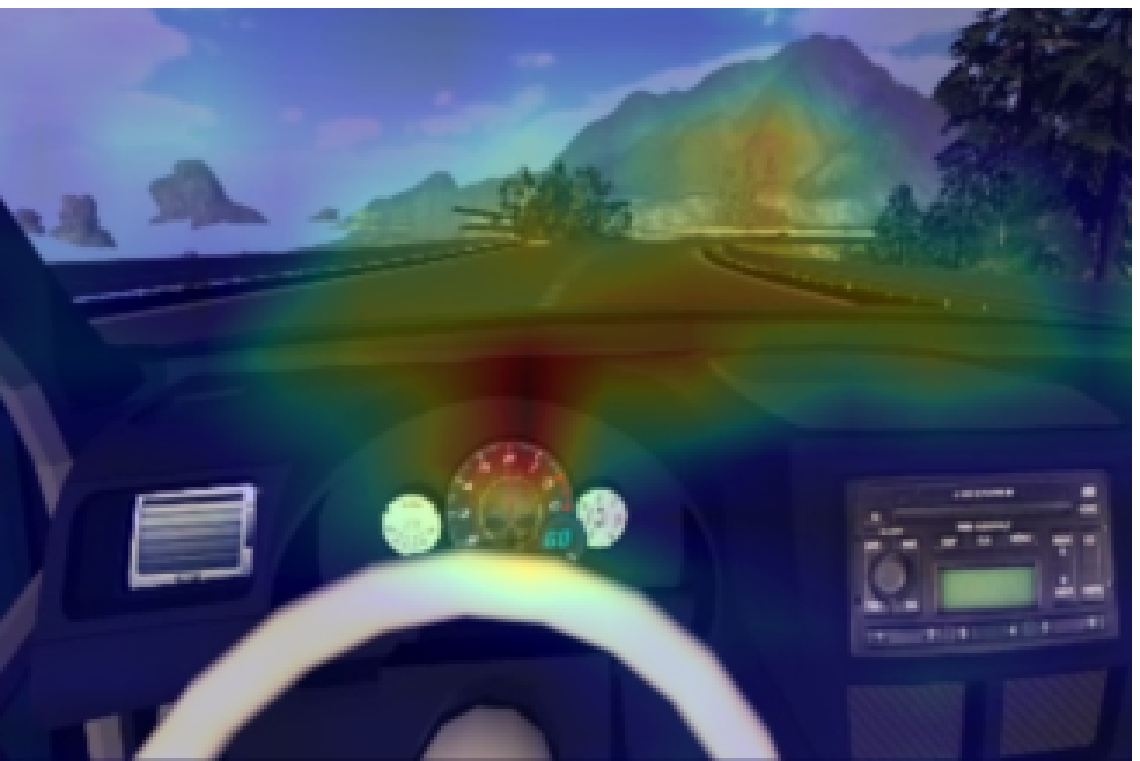}\hfill
    \includegraphics[width=.3\linewidth]{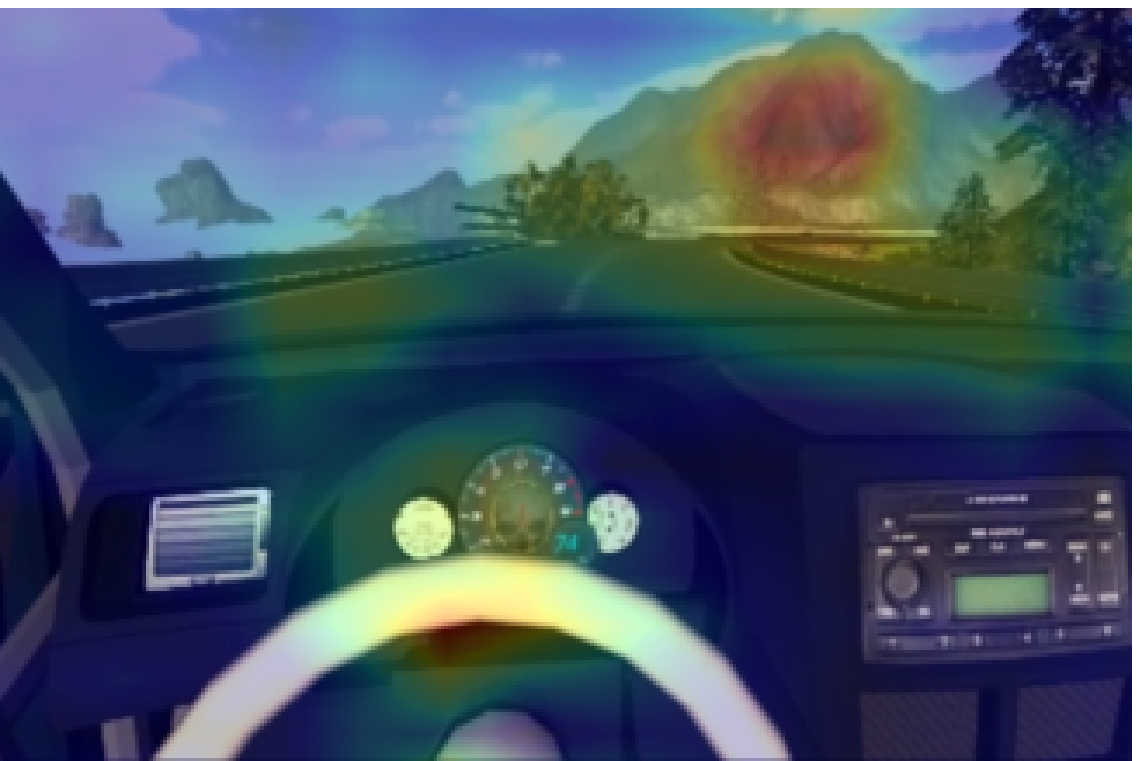}\hfill
    \includegraphics[width=.3\linewidth]{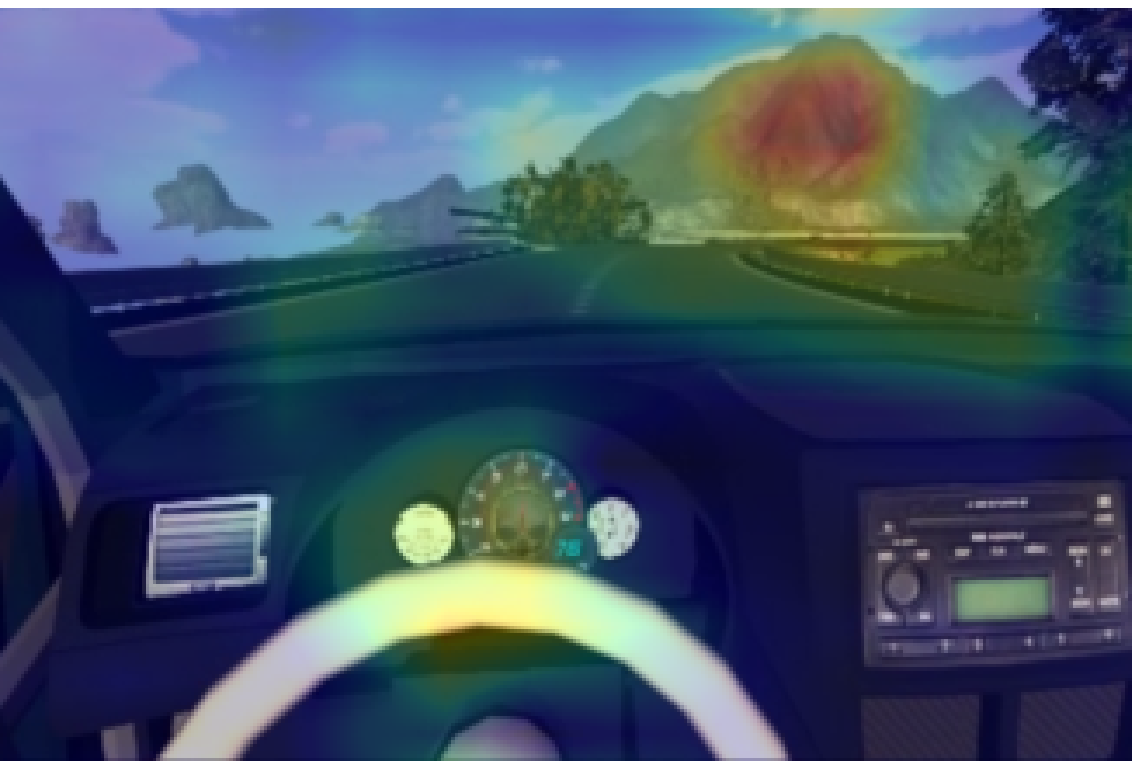}
    \caption{Turning right (fall off a cliff) in Period 1}
    \label{sfig:rightin period5}
    \end{subfigure}

\caption{Attention maps from Grad-CAM at equally sampled time points (left =start; right=end) for Periods 1 and 5. In Period 5, which is the closest period to the final decision, the DNN focuses on tree or cliff. In the earlier Period 1, the DNN puts more focus on the steering wheel or the ridge of the hill.}
\label{fig:grad_cam_result}
\vspace{-0.5cm}
\end{figure}

As shown in Figure \ref{fig:grad_cam_result} (a), (b) for an example segment of Period 5, both attention maps focus on the central, steering wheel part. Beyond this, however, there are crucial differences in the resulting attention map depending on the condition. In the collision condition (Figure \ref{fig:grad_cam_result}(a)), the network fixates areas nearby the tree in every frame of the period. In contrast, the model concentrates on both tree and cliff in the fall condition (Figure \ref{fig:grad_cam_result}(b)), but as time progresses, the concentration on the cliff becomes more prominent. 

The visualizations for the earliest Period 1 showed that the network focuses on the steering wheel and hill ridges in the collision condition and on similar areas on both trees and cliffs in fall conditions (cf. Figure \ref{fig:grad_cam_result} (c), (d)).

These qualitative differences were highly consistent in all input videos, indicating that the DNN is paying attention to meaningful - and expected - visual features for making the prediction. 

\subsection{Effect of spatial degradation}
We next test the generalizability and robustness of the model by degrading the visual input. We first add spatial Gaussian noise to an image during testing either in the top 60\% or the bottom 40\% of the image.

Performance in the two conditions for all periods is shown in Figure \ref{fig:Various_experimental}(a). Overall, bottom blurring has virtually no effect on performance, whereas top blurring significantly reduces performance in the initial period, and especially in the final period (performance in Period 5: $72.76\%$ for top-blurring versus $81.85\%$ for bottom blurring and $81.97\%$ for unblurred images). Hence, as one would expect from the attention-map analysis in Figure \ref{fig:grad_cam_result}, top-blurring considerably reduces the networks ability for predicting decision-making.

\begin{figure}[tp!]
    \begin{subfigure}[h]{\textwidth}
    \includegraphics[width=.3\linewidth]{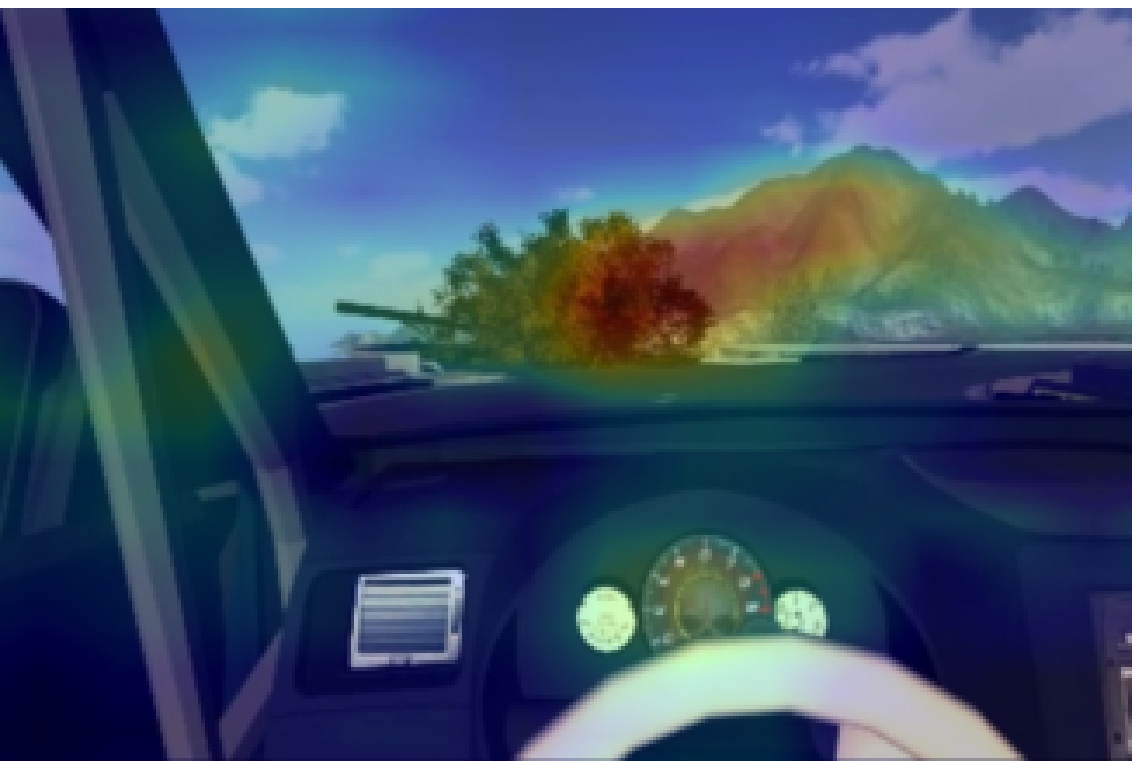}\hfill
    \includegraphics[width=.3\linewidth]{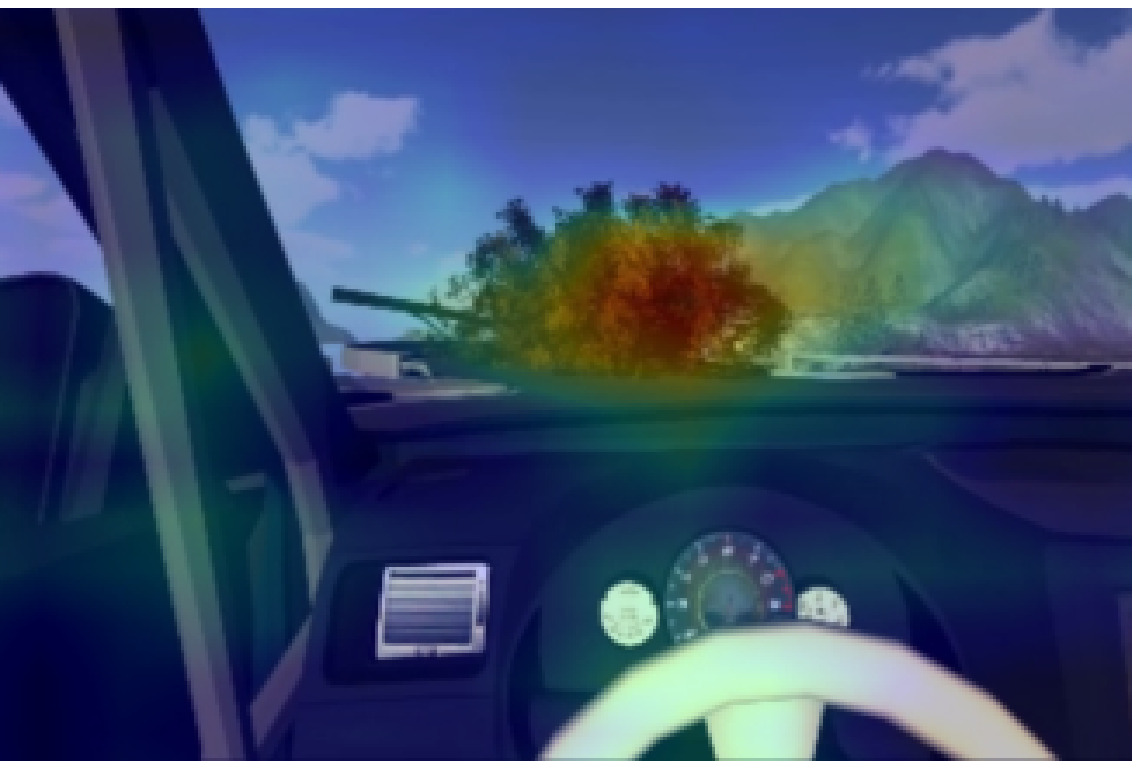}\hfill
    \includegraphics[width=.3\linewidth]{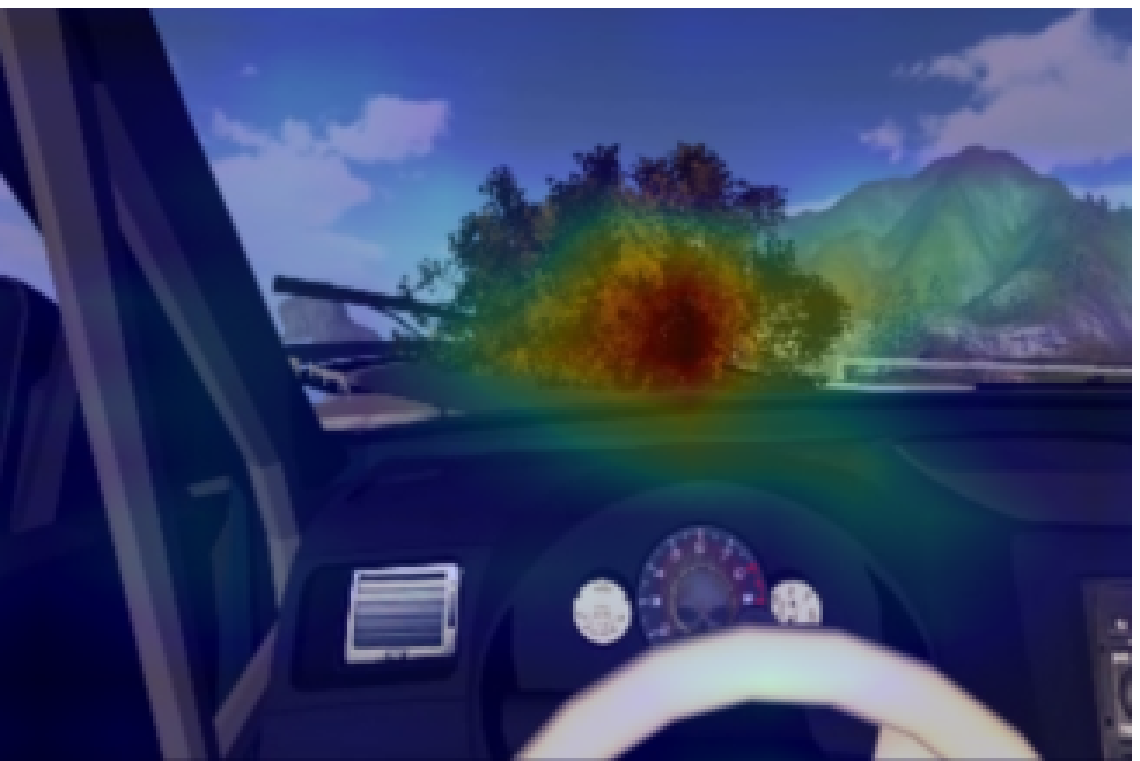}
    \caption{Original frame}
    \label{fig:origin}
    \end{subfigure}
    \\[\smallskipamount]
    \begin{subfigure}{\textwidth}
    \includegraphics[width=.3\linewidth]{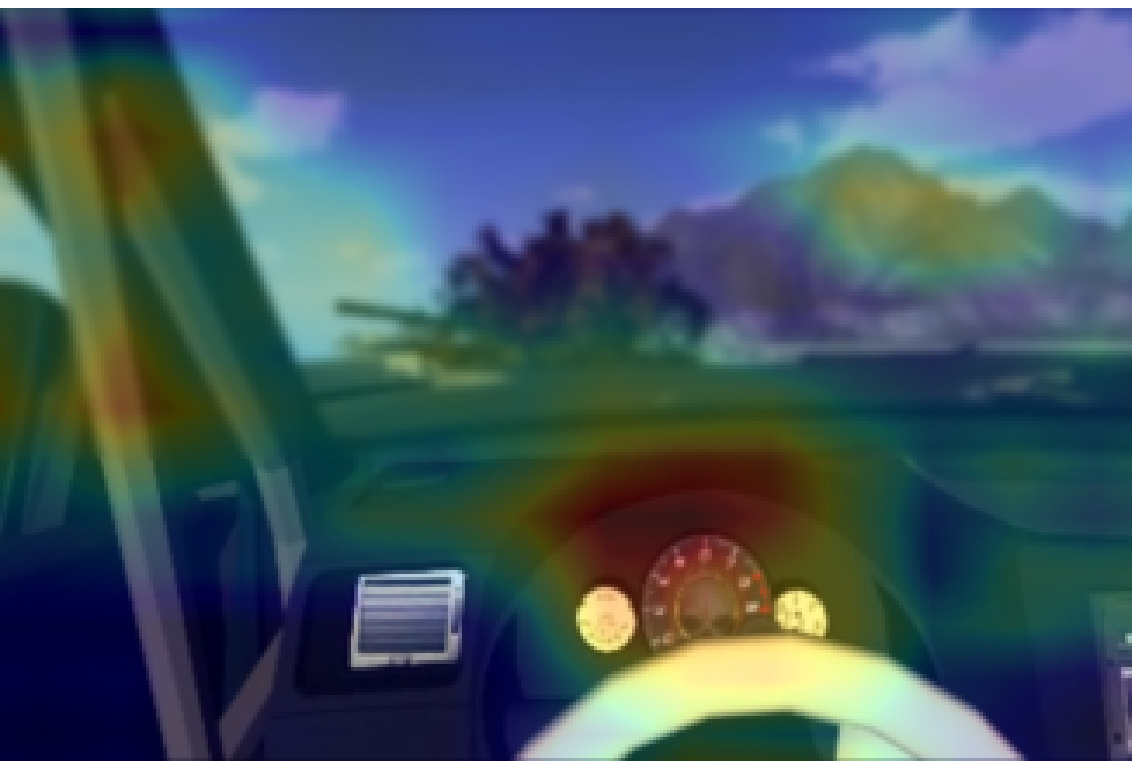}\hfill
    \includegraphics[width=.3\linewidth]{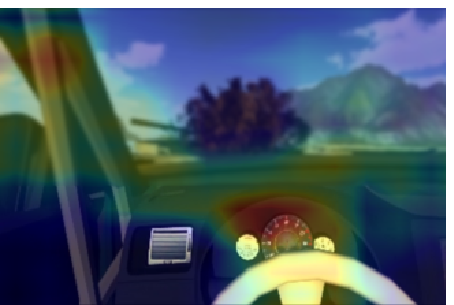}\hfill
    \includegraphics[width=.3\linewidth]{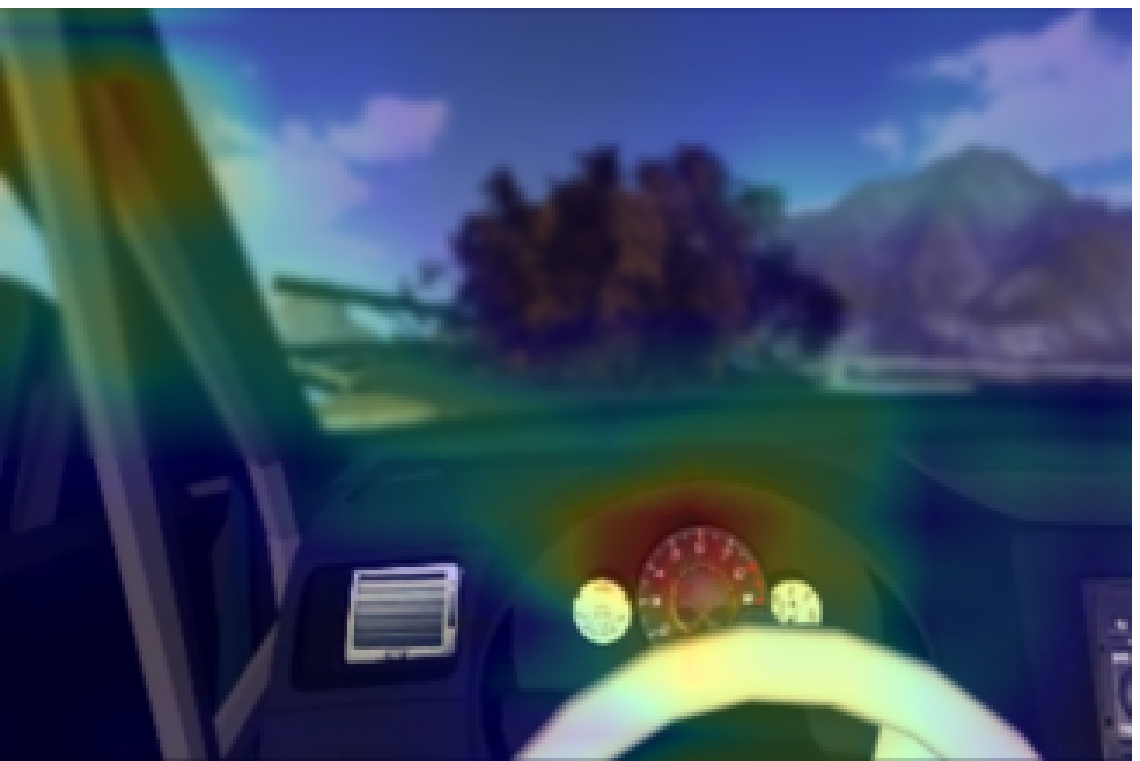}
    \caption{Top 60\% blurring}
    \label{fig:top_blur}
    \end{subfigure}
    \\[\smallskipamount]
    \begin{subfigure}{\textwidth}
    \includegraphics[width=.3\linewidth]{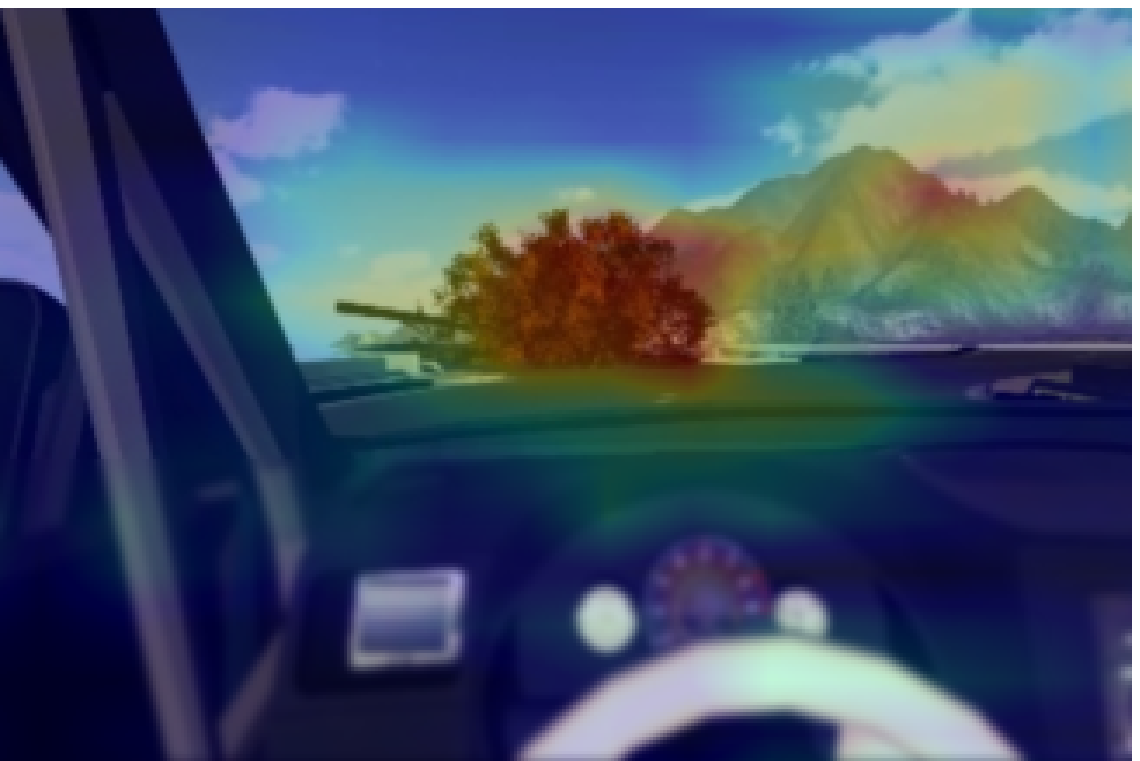}\hfill
    \includegraphics[width=.3\linewidth]{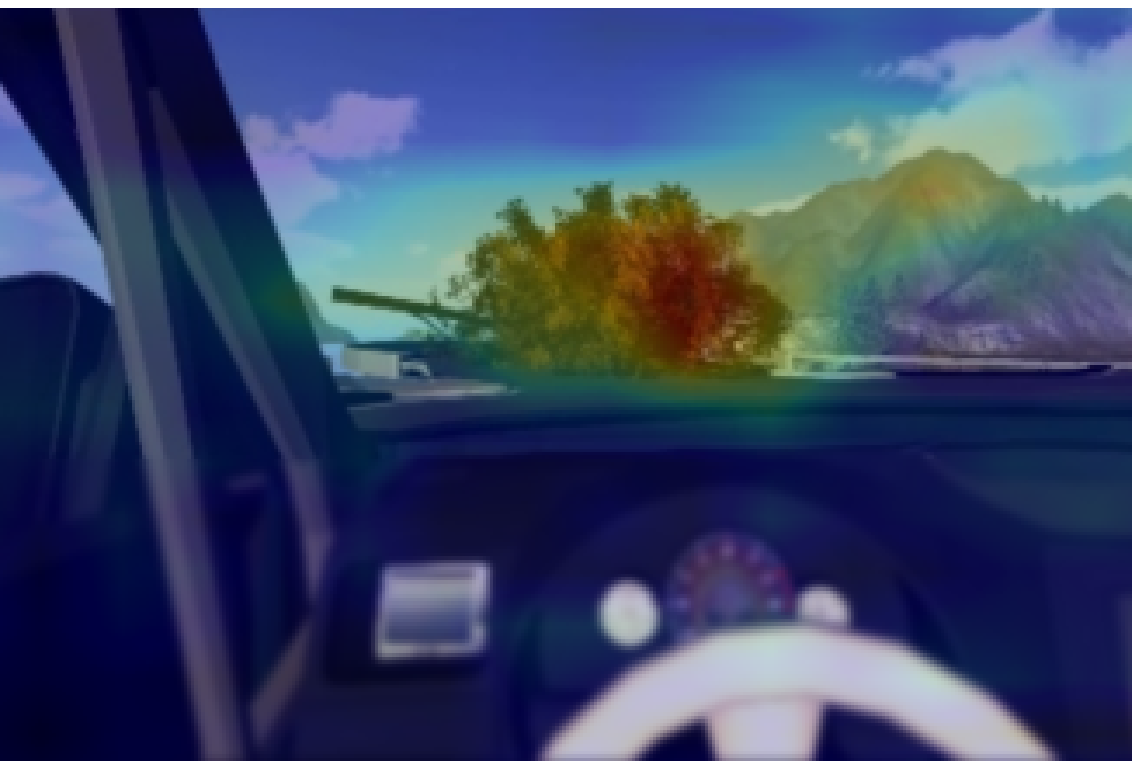}\hfill
    \includegraphics[width=.3\linewidth]{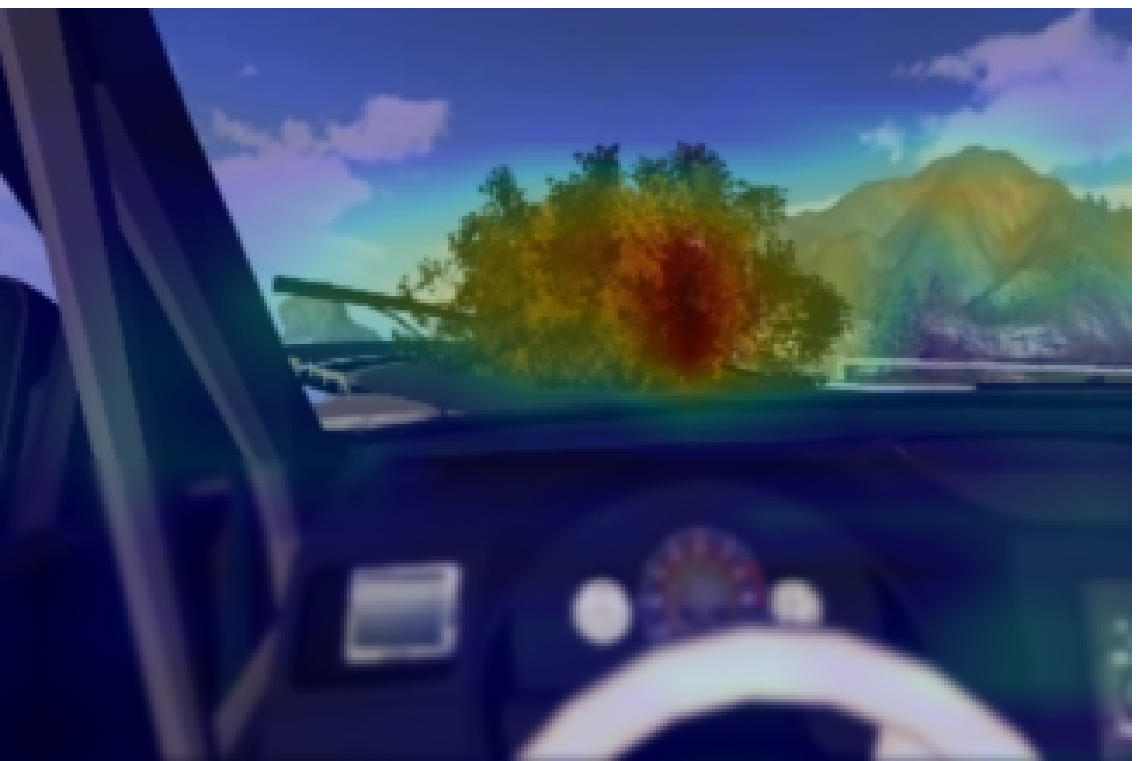}
    \caption{Bottom 40\% blurring}
    \label{fig:bottm_blur}
    \end{subfigure}
\caption{Effect of spatially-selective blurring on discriminative features. From left to right: start frame to end frame of Period 5. The top row shows discriminative features for the original sequence. Blurring effects are shown applied to the top 60\% of the original frame (middle row), or the bottom 40\% (bottom row).}
\label{fig:blurring_result}
\vspace{-0.5cm}
\end{figure}

Figure \ref{fig:blurring_result} compares the attention maps of the non-blurred with the two blurred conditions to confirm the accuracy results. Indeed, when blurring the top part, almost all activation focuses on the bottom, non-blurred input, virtually disregarding the important features outside the car. Perhaps the remaining focus on the steering wheel may lead to the above-chance prediction performance that  could still be observed. As one would expect, blurring the bottom part of the image has virtually no effect on the attention map, barring a slight decrease of focus on the steering wheel (see Figure \ref{fig:blurring_result}(c)). 

\subsection{Temporal analysis}
In general, continuous sequences of frames in the video are known to be critical for understanding events (see, for example,\cite{cunningham2009dynamic} for a detailed study on the dynamics of facial expressions). To determine the importance of different temporal aspects, we next present experiments that modified the number of input frames or changed the temporal order of frames.

\subsubsection{Changing number of frames:}
\begin{figure}[htpb!]
     \centering
     \begin{subfigure}[t]{0.8\textwidth}
         \centering
         \includegraphics[width=.99\linewidth]{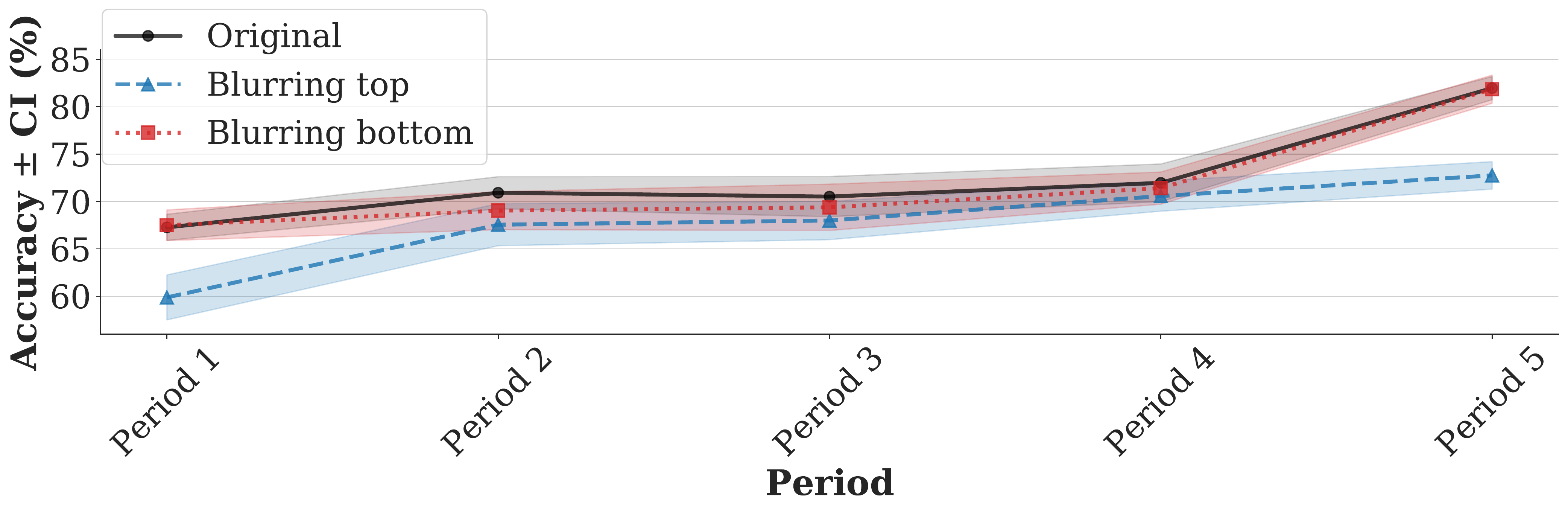}
         \caption{Spatially-selective blurring}
         \label{fig:Blur}
     \end{subfigure}
     \begin{subfigure}[t]{0.8\textwidth}
         \centering
         \includegraphics[width=.99\linewidth]{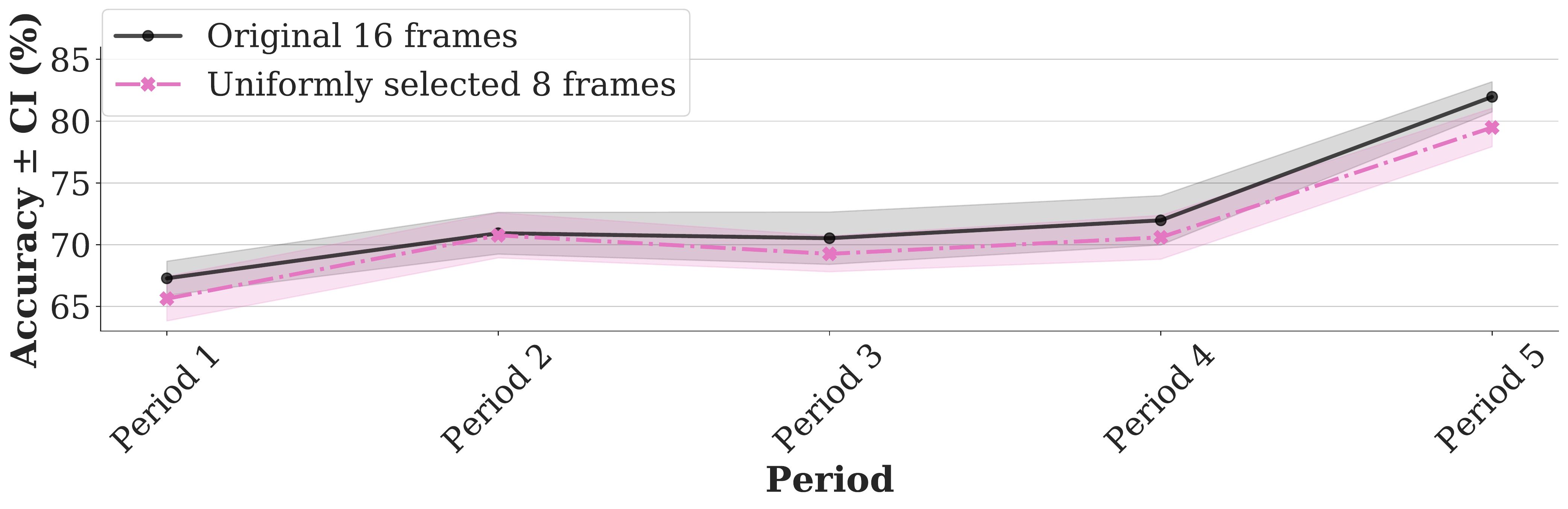}
         \caption{16 frames versus subsampled 8 frames}
         \label{fig:figure_16_8_frame}
     \end{subfigure}
    \begin{subfigure}[t]{0.8\textwidth}
         \centering         
         \includegraphics[width=.99\linewidth]{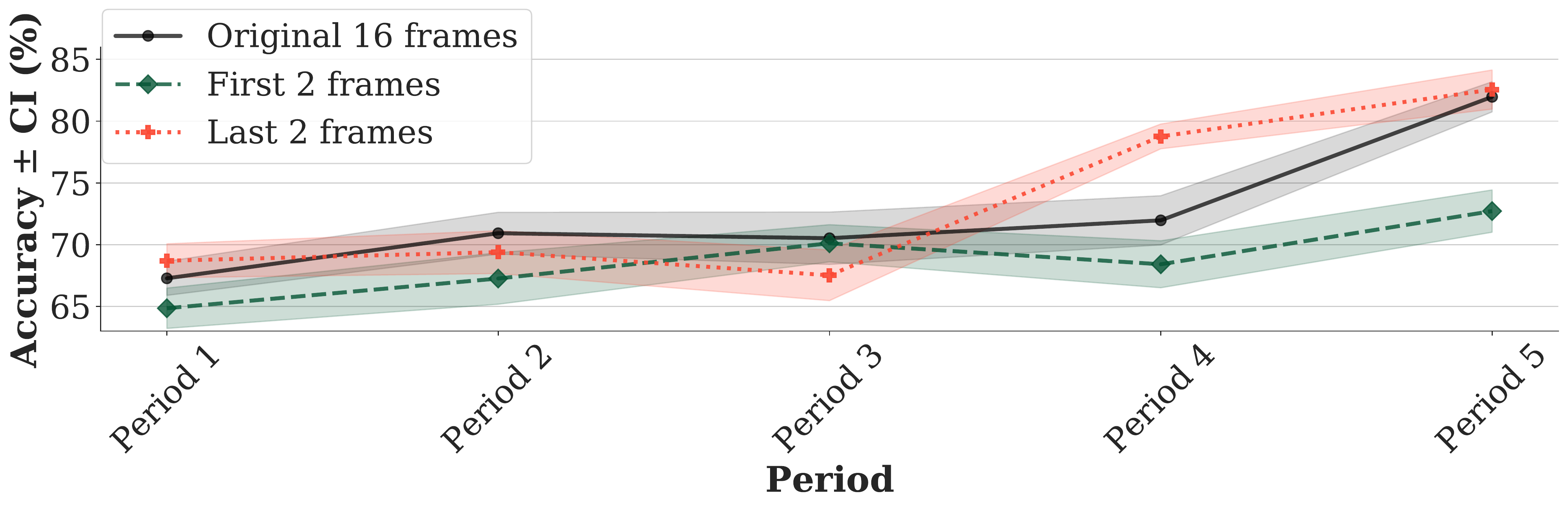}
         \caption{First 2 frames versus the last 2 frames}
         \label{fig:figure_firstlast}
     \end{subfigure}    
    \begin{subfigure}[t]{0.8\textwidth}
         \centering
         \includegraphics[width=.99\linewidth]{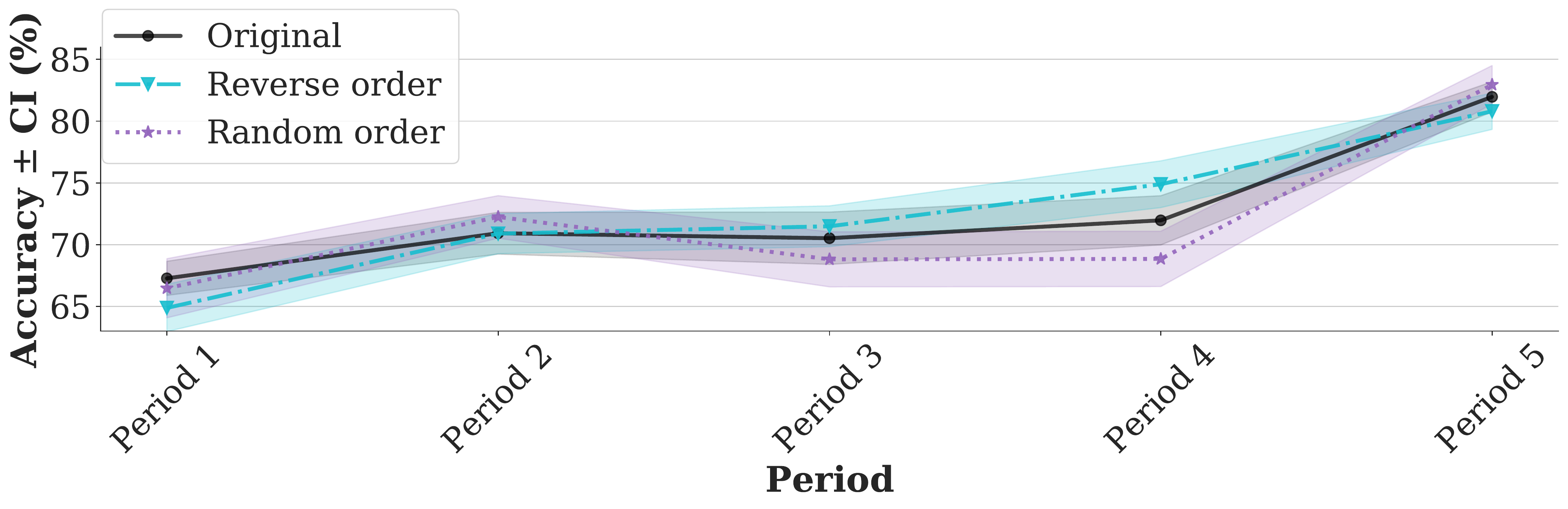}
         \caption{Random order versus reversed order} 
         \label{fig:d}
     \end{subfigure}    
\caption{Prediction accuracy for manipulations in space (blurring (a)) and time (16 vs 8 frames (b), first vs last frames (c), shuffling and time-reversal (d)).}
\label{fig:Various_experimental}
\end{figure}

Reducing the number of input frames from 16 frames to 8 frames showed only minor decreases in performance (Figure \ref{fig:Various_experimental} (b)). A further reduction from the original 16 frames to only 2 frames (Figure \ref{fig:Various_experimental} (c)) showed varying results: original performance levels could only be obtained with the last 2 frames of the segment, whereas using the first 2 frames of the Period resulted in an overall drop in performance, especially towards the final Period. Interestingly, for Period 4, prediction accuracy for the final 2 frames outperformed those obtained by all 16 frames, reaching almost peak accuracy. \vspace{-0.3cm}

\vspace{-0.5cm}

Overall, these results seem to suggest that prediction seems to rely on the final, few frames of each period with limited advantages of adding further frames for temporal context.\vspace{-0.5cm}

\subsubsection{Changing temporal order:}
Given prior results from human studies on the importance of the direction of time and the preservation of temporal structure in general \cite{cunningham2009dynamic}, we next reversed time or shuffled the frames. As Figure \ref{fig:Various_experimental} (d) shows, this has very little effect on performance, indicating that temporal structure itself bears little importance for the model.\vspace{-0.2cm}

\section{Conclusion}
\vspace{-0.2cm}
In this paper, we investigated the difference between humans and a DNN in predicting the final decision in an accident situation. Our results showed that both humans and DNNs increased in accuracy as the time period approached the actual decision - a result in line with expectations, as both the path of the car and the steering movements may ``settle" on their final direction at that point. We also found that the DNN made more accurate judgments compared to the human at early time points. Grad-CAM analysis further showed that its attended features are meaningful in terms of semantic content: interestingly, for the early time period, in which the DNN outperforms humans, its focus is on a wider view of the scene (including the ridge of hill) as well as the steering wheel, whereas in later time periods it focuses on closer things (tree or cliff). In future work, we will compare these computational attention maps to those obtained with further human experiments, using, for example, eye-tracking. 

Moreover, several analyses were conducted to dive deeper into the underlying spatial and temporal features that may give rise to the prediction accuracy. In terms of spatial features, blurring showed that the outside, top view drove most of the recognition performance, which matches with expectations.  In terms of temporal features, we found that the number of input frames does affect performance to some degree, but also showed that even two frames - when properly selected - still yield high performance. Again, it remains to be seen whether human performance would be similar with the same kind of input reduction. 

Although most of the analyses so far would match with qualitative expectations about the important features for predictions, our analysis of time reversal and shuffling indicates no adverse effects of these manipulations. This is perhaps somewhat surprising given ample evidence in the human literature that time structure is crucially important in event analysis \cite{cunningham2009dynamic,liu2019human}. Here, human experiments would most likely yield quite reduced performance, indicating that especially the learned \textit{temporal} representation of DNNs may be different to those of humans. Further experiments will be necessary with different pre-training schemes and ``deeper" visual hierarchies (vision transformers \cite{dosovitskiy2020image}) to investigate in more detail to what degree human and DNNs representations are similar.

Finally, we want to note that comparing data from human and computational experiments is bound to be complex: the DNN was specifically trained on a decision that was tested later, hence, there was no notion of task generalizability; similarly, humans will attach ``meaning" to the outcome of the decision, with the left/right choice carrying consequences that are implicitly understood - such semantic grounding is so far missing from the DNN representation. It will be interesting to see how the current trend of research towards ``foundation models"\cite{bommasani2021opportunities} will create frameworks that are capable of producing more human-level, semantically-rich, and generalizable task solutions.   

\vspace{-0.5cm}
\subsubsection{Acknowledgements:}
{\small This work was supported by the National Research Foundation of Korea under Grant NRF-2017M3C7A1041824 and by two Institute of Information and Communications Technology Planning and Evaluation (IITP) grants funded by the Korean government (MSIT): Development of BCI based Brain and Cognitive Computing Technology for Recognizing User’s Intentions using Deep Learning (2017-0-00451), and Artificial Intelligence
Graduate School Program (Korea University) (2019-0-00079).}
%
%
%
\vspace{-0.6cm}
{\small\bibliography{references}}
%
%
%
\end{document}